\documentclass[letterpaper, 10 pt, conference]{ieeeconf}  % Comment this line out if you need a4paper

\IEEEoverridecommandlockouts                              % This command is only needed if 
\usepackage{multirow}
\usepackage{graphicx}
\usepackage{etoolbox}
\usepackage{hyperref}
\hypersetup{hidelinks,
            colorlinks=true,
            allcolors=black,
            pdfstartview=Fit,
            breaklinks=true}
\makeatletter
\patchcmd{\@makecaption}
  {\scshape}
  {}
  {}
  {}
\makeatletter
\patchcmd{\@makecaption}
  {\\}
  {.\ }
  {}
  {}
\makeatother

\usepackage[table,xcdraw]{xcolor}
\title{\LARGE \bf
LEBP --- Language Expectation \& Binding Policy: A Two-Stream Framework for Embodied Vision-and-Language Interaction Task Learning Agents}
\author{Haoyu Liu$^{1}$, Yang Liu$^{2}$, Hongkai He$^{1}$ and Hangfang Yang$^{1,*}$ % <-this % stops a space
\thanks{* Corresponding author.}% <-this % stops a space
\thanks{$^{1}$ Center for Applied Statistics, School of Statistics, Renmin University of China, Beijing 100872, China. 2017201665@ruc.edu.cn (H.L.);  2017201609@ruc.edu.cn(H.K.); hyang@ruc.edu.cn(H.Y.).}%
\thanks{$^{2}$ Samsung Research China Beijing (SRCB). Beijing 100028, China. yang9004.liu@samsung.com(Y.L.).
}}

\begin{document}

\maketitle
\thispagestyle{empty}
\pagestyle{empty}

%%%%%%%%%%%%%%%%%%%%%%%%%%%%%%%%%%%%%%%%%%%%%%%%%%%%%%%%%%%%%%%%%%%%%%%%%%%%%%%%
\begin{abstract}
People always desire an embodied agent that can perform a task by understanding language instruction. Moreover, they also want to monitor and expect agents to understand commands the way they expected. But, how to build such an embodied agent is still unclear. Recently, people can explore this problem with the Vision-and-Language Interaction benchmark ALFRED, which requires an agent to perform complicated daily household tasks following natural language instructions in unseen scenes. In this paper, we propose LEBP – Language Expectation with Binding Policy Module to tackle the ALFRED. The LEBP contains a two-stream process: 1) it first conducts a language expectation module to generate an \textbf{expectation} describing how to perform tasks by understanding the language instruction. The expectation consists of a sequence of sub-steps for the task (e.g., Pick an apple). The expectation allows people to access and check the understanding results of instructions before the agent takes actual actions, in case the task might go wrong. 2) Then, it uses the binding policy module to \textbf{bind} sub-steps in expectation to actual actions to specific scenarios. Actual actions include navigation and object manipulation. Experimental results suggest our approach achieves comparable performance to currently published SOTA methods and can avoid large decay from seen scenarios to unseen scenarios.
\end{abstract}

%%%%%%%%%%%%%%%%%%%%%%%%%%%%%%%%%%%%%%%%%%%%%%%%%%%%%%%%%%%%%%%%%%%%%%%%%%%%%%%%
\section{Introduction}
People have always wanted an agent to understand and execute the instructions given by natural language. More importantly, people also want to monitor and expect agents to understand their commands the way they think, such as \textit{checking if the agent picks the right cup for coffee}. This is crucial in human-agent interaction scenarios, such as housekeeper robots \cite{shridhar2020alfred}, virtual assistants\cite{szlam2019build}, etc. While, how to build such an embodied agent remains an open problem. Recently, people have proposed a vision-and-language interaction benchmark ALFRED, which we can benefit from exploring this problem. ALFRED  proposes a new task that requires an agent to perform complicated daily household tasks following natural language instructions in a virtual interactive environment \cite{ai2thor}. To complete a task, agents need to understand and ground language into the scene, navigate (e.g., Move and Turn) and manipulate an object (e.g., Pick, Put, and Toggle).

Moreover, the agent needs to be evaluated in an unseen scenario instead of scenarios where it is trained. Currently, published methods attempt to solve this with Seq2Seq methodology  \cite{sutskever2014sequence}, mainly taking language instructions and vision frames (RGB images) as input sequences and directly mapping them into output action sequences to execute tasks. Further improvements either focus on applying advanced methods like BERT \cite{devlin2018bert}, designing modular multi-modal modules for better state capturing \cite{Embert} or proposing episodic modules for a transformer to catch key memory frames  \cite{ET}. 

Although these approaches show clear progress, there are still issues that cause our concern. First, the Seq2Seq methodology may be difficult to overcome from the diversity of object texture, agent location, and camera angle of the scene. Second, before the agent performs actual actions, it is unclear how to check if the agent understands the instruction correctly as people expected until it makes a mistake. This is unacceptable for human-robot interaction.

\begin{figure}[]
  \centering
  
  \includegraphics[width=0.47\textwidth]{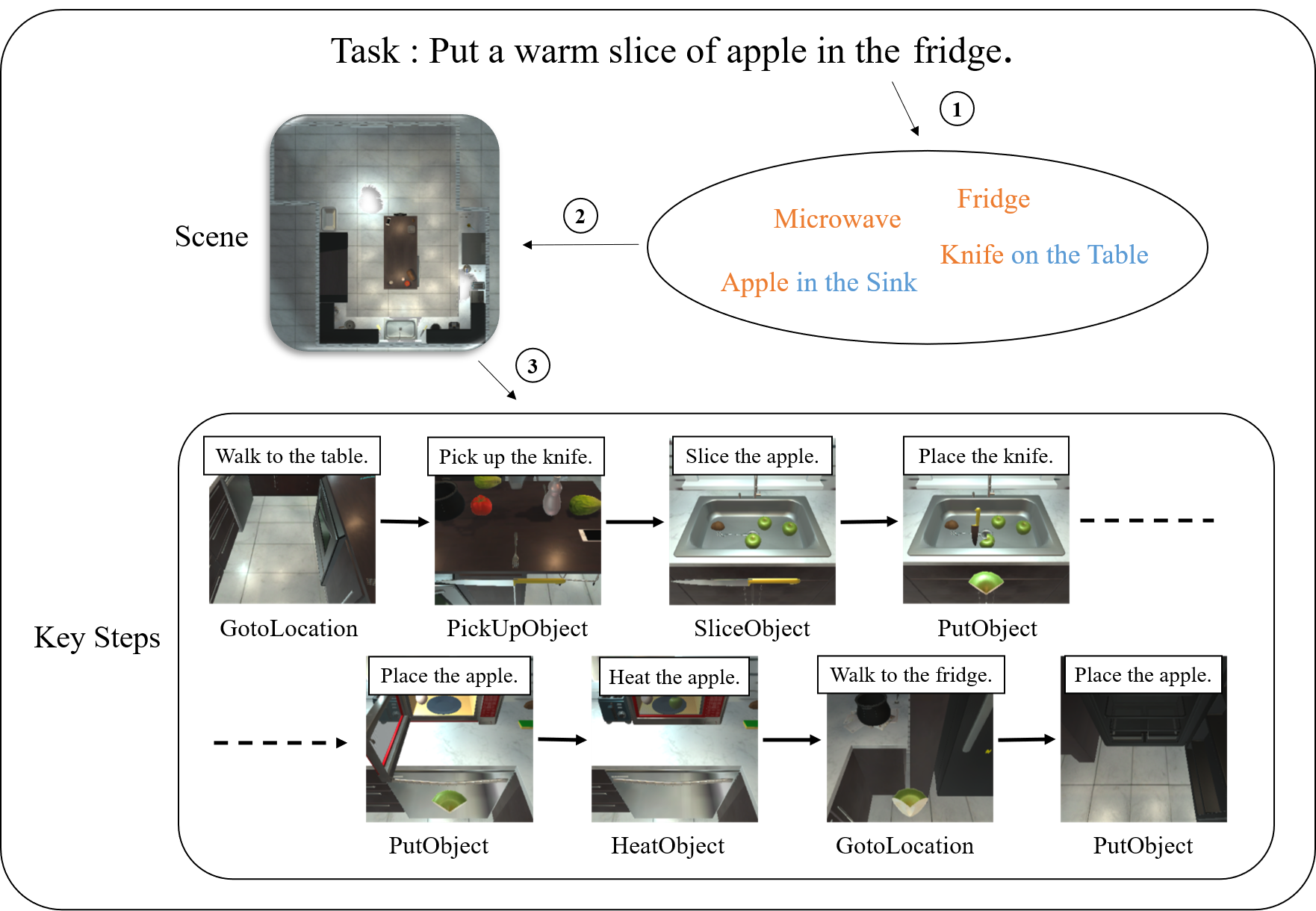}
  \caption{Demonstration of human task planning and execution process. When a person receives an instruction, he does not immediately start the actual action, but, based on the instruction, generates an expectation of how to perform the task, which usually consists of several sub-steps. For example, Task: \textit{Put a warm slice of apple in the fridge} can be divided into sub-steps including from \textit{Walk to the table} to \textit{Place the apple}. He can discuss it with people to understand his expectations of task and then perform the task. We borrowed this idea to implement our vision-language-interaction model. The expectations (i.e. key steps) generated by the agent based on language instructions can be used to check the correctness of its planning to improve human-robot interaction.
  }
  \label{mechanism}
      
\end{figure}

In this paper, we propose a two-stream framework: Language Expectation \& Binding Policy (LEBP) to separate the language understanding and action execution in an embodied agent. Our approach is inspired by the human process in understanding language and executing tasks (see Fig\ref{mechanism}). Instead of performing actions immediately, our agent model first conducts a language expectation module to generate an \textbf{expectation} describing how to perform tasks by understanding the language instruction. The expectation consists of a sequence of sub-steps for the task (e.g., Pick an apple). The expectation allows people to access and check the understanding results of instructions before the agent takes actual actions, in case the task might go wrong. Then, we use the binding policy module to \textbf{bind} sub-steps in expectation to actual actions to specific scenarios. Actual actions include navigation and object manipulation.

%First, LEBP contains a language expectation generation module that can generate a sequence of sub-steps that describe the expectation of how the agent will perform the whole task. For example, given a task \emph{put an arm slice of apple in the fridge}, the sub-step would be \emph{take a knife}, \emph{slice the apple}. During expectation generation, in the process of generating expectations, we also consider the specific information of the scene. The results of language expectation which reflects the understanding of language instruction can be used for visualization. Thus, people can check the schedule.

To evaluate our approach, we test it on the ALFRED benchmark. Experimental results suggest our approach achieves comparable performance to currently published SOTA methods. The results also suggest that LEBP can alleviate the potential overfitting of the strong correlation between language instructions and visual detail (e.g., RGB, texture, etc.) of scenarios from training trajectories. Thus, our method can avoid large decay from seen scenarios to unseen scenarios. Moreover, LEBP is more conducive to directly injecting environment prior information and introducing spatial structure. Therefore, since LEBP can improve interactivity via language expectation and meanwhile the overall performance, we argue that it provides reasonable insights for the future solution of this task.

\section{Related Work}

%\subsection{Embodied Language Understanding and Interaction}

In the past two years, Embodied AI has attracted more and more attention\footnote{https://embodied-ai.org/} \cite{NIPS2016,embodiedqa,RoomR,shen2021igibson,8954627}. Among Embodied AI tracks, Vision-and-Language Interaction is a very challenging problem. Recently, people have proposed ALFRED to explore this direction \cite{shridhar2020alfred}, which requires language and vision understanding, interaction mode design. Moreover, it is a preview of the in-house agent. In ALFRED, the agent is asked to perform corrective actions to complete the task described in language instruction by navigation and manipulation. The baseline system for the ALFRED benchmark employs a Seq2Seq framework \cite{shridhar2020alfred} containing an attention module and progress monitoring, mapping the goal-instruction language and visual observation into actions. Most further studies follow this paradigm and make progress via introducing a pre-trained vision module to generate object masks \cite{LAV}, designing a modular architecture to independently predict objects and actions \cite{Moca}, or applying an advanced transformer-based model \cite{Embert, ET, Hitut}. However, \cite{ABP} suggests that a long sequence in inputs and trivial actions in outputs increase the difficulty for the Seq2Seq model to capture the state of agents and the risk of over-fitting in training scenarios is rather high. To assist the process of a long task, \cite{HLSM} predicts the sub-goal of the current state and utilizes it as auxiliary information for further task execution. 

Meanwhile, the core idea of these works seems to ignore the basic needs of human-agent interaction: interactivity and interpretability. Specifically, in a real-world application, humans need to understand and monitor the behavior of the autonomous agent in case it performs the task in the wrong way. However, current methodology support such demands badly, since prevailing end-to-end neural-based architecture (e.g., reinforcement learning \cite{reinforcement}, Seq2Seq \cite{sutskever2014sequence}) always work in black-box style and treats task as one-time pass. Then, we believe it is necessary to introduce structures to improve interactivity and interpretability.

\begin{figure*}[!htbp]
\centering
\includegraphics[scale=0.53]{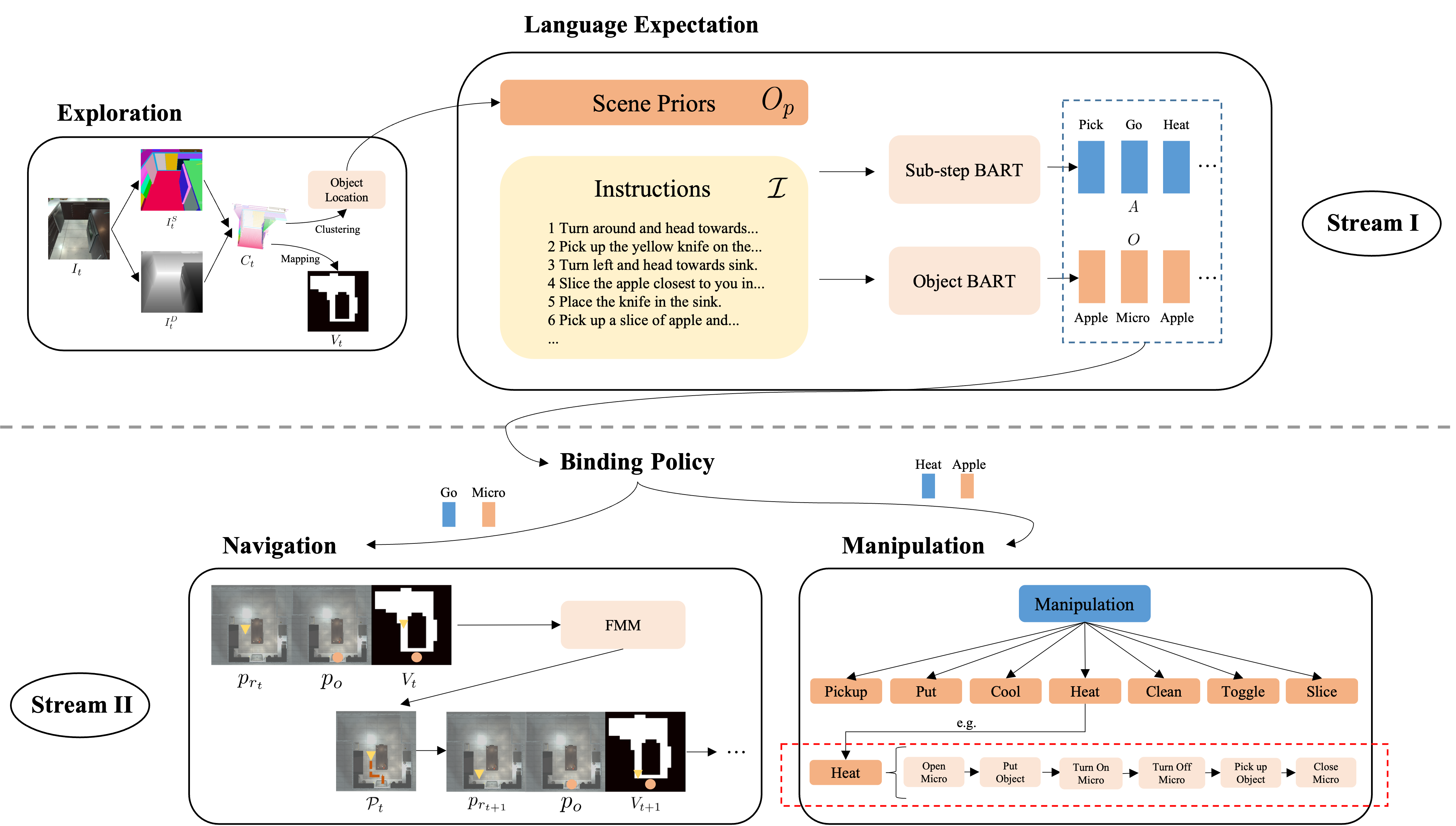}
\caption{Overview of the LEBP framework. LEBP consists of a two-stream framework where stream I) the language expectation module decouples the whole task into sub-steps and II) the binding policy part successively executes sub-steps following language expectation sequences. The sub-step is summarised as two types: navigation and object manipulation. For navigation, a 2D obstacle map $V_t$ is constructed and updated dynamically in the exploration process with a semantic point cloud generated from the depth and segmentation frames, which are predicted by learned modules. Meanwhile, point cloud in specific object labels is clustered to get the center coordinate, which is memorized as object location. The manipulation part designs a corresponding set of basic actions to execute specific manipulation steps, just as Manipulation Task Heat (red box) shows.} 
\label{Model}
\end{figure*}

\section{Method}

LEBP utilizes a two-stream framework shown in Figure \ref{Model}, containing language expectation (stream I) and binding policy (stream II). In stream (I): language expectation module generate sub-steps of expectation based on the whole task described in language instruction. With step-by-step instructions, this module injects the scene prior knowledge which is induced through environment exploration into the Seq2Seq model called BART \cite{bart} to generate the sub-step sequence and object sequence related to each sub-step. For example, object \emph{Apple} related to sub-step \emph{PickUpObject}. Stream (II) refers to the binding policy in LEBP, which executes each sub-step with actual actions including navigation and manipulation. For navigation, we construct the scene map dynamically from initial exploration and plans the route for the agent to reach locations. For object manipulation, the binding policy module arranges the corresponding set of basic actions according to the manipulation type (e.g. the Heat manipulation consists of six atomic actions: Open, Close, PickUp, Put, ToggleOn, and ToggleOff). The object sequence here gives a specific needed object in each sub-step, indexing object location for navigation and pixel-level mask for manipulation. Thus, we can complete a task step-by-step.

\subsection{Language Expectation}

As mentioned above, we use the language expectation module to generate expectation, where we desire agents the ability to understand instructions and plan steps in a specific environment to break down the task. The language expectation module utilizes the BART structure to handle two Seq2Seq tasks: sub-step sequence and object sequence, taking step-by-step language instruction as the input. Specifically, We first concatenate the step-by-step instructions and derive token embedding $\mathcal{L}$. 

To adapt the instructions to the environment, we extract the scene prior information via an embedding vector $\mathcal{O}_p = \{o_1,o_2,...,o_{n_p}\}$, where $o_i$ represents the object embedding vector in the current scene \footnote{There are 120 different scenes in the ALFRED benchmark and each scene is a household room containing a different set of objects.} after the exploration. Thus, the prior scene information serves as an auxiliary embedding to help avoid planning nonexistent objects. Then we join the word tokens embedding $\mathcal{L}$ and scene embedding $\mathcal{O}_p$ together and feed it into the BART module to finish the Seq2Seq task:

$$A = {\rm BART_1} ([\mathcal{L}; \mathcal{O}_p])$$
$$O = {\rm BART_2} ([\mathcal{L}; \mathcal{O}_p])$$

where $[; ]$ denotes concatenation, $A = \{ a_1, a_2, ..., a_{l}\}$ denotes the sub-step sequence, and $O = \{ o_1, o_2, ..., o_{l}\}$ denotes the object sequence.\\

The language expectation module performs an important role in the whole framework. It breaks down the whole task and enables the agent to understand instructions and plan steps, which provides a foresee of what agents will do to enhance the human-agent manipulation ability.

\subsection{Map Construction and Scene Exploration}

After breaking down the whole task into sequences, we tend to design modular architectures to realize corresponding sub-step functions: navigation and object manipulation. 

We believe that when the agent enters an unknown scene, it needs to first find the object to be operated, on and then execute tasks e.g. cleaning or heating the object. These operations all need a preliminary exploration of the environment, such as obtaining the approximate location of objects and finding the location of possible obstacles in the room, so as to avoid obstacles in the implementation of executing tasks. In this work, we construct a 2D navigation map and propose an initial exploration process in the scene to detect objects and memory their locations.

\subsubsection{Map Construction}
In general, the agent will build up the 2D map $V_t$ dynamically in the process of exploration to help it better understand the scene and execute the next tasks. Specifically, the interval of $V_t$ is 0.25, matching the grid length in the simulator. Each value in $V_t$ is $0/1$ to represent obstacles and the size of the map equals the room size. 
At each time $t$, the agent egocentric RGB observation $I_t$ is passed into a learned semantic segmentation model and depth model to predict pixel-level semantic label $I^S_t$ and depth frame $I^D_t$. The predicted depth frame is processed to generate the point cloud \footnote{the point cloud is transformed by the camera extrinsic parameters, which can be found in Ai2thor \cite{ai2thor} webset.} $C_t=\{p_1, p_2, ..., p_N\}$, where each point $p_i \in C_t$ contains a 3D coordinate $(x_i,y_i,h_i)$ and a semantic label $o_i$.  
Then each value in map $V_t$ is calculated as:

$$V_t(i,j)=\mathbf{E}( \sum_k \mathbf{E}(0<h_k<H))$$

where $H$ denotes the height of the agent and $\mathbf{E}$ denotes indicator function projecting height to 0-1 discrete data. This means if all points at $(i,j)$ are higher than the agent, it values $0$ and represents a passable grid. For each observed object category $o$, we cluster its points' coordinates by the DBSCAN \cite{DBSCAN} algorithm and memory the cluster center as its location $p_o$.

\subsubsection{Scene Exploration}

Initial exploration in the scene is adapted when performing new tasks. In the beginning, we spilt the coordinates in the 2D map $V_t$ into two sets: the explored set $P_s$ and unknown set $P_u$. We constantly sample coordinates in $P_u$ and explore the unknown area, updating the 2D map and the object location memory concurrently at each time step. The exploration ends when all coordinates in the unknown set are explored (i.e. the coordinate is seen by the agent). After exploration, the map and location memory is relatively complete for upcoming binding policy execution.

\subsection{Binding Policy}

The binding policy is well designed for sub-step execution summarised as two categories: navigation and object manipulation. In the whole LEBP framework, the binding policy serves as the down-streaming modular parts for the language expectation module, whose relationship is like an agent's limb and brain.

\subsubsection{Navigation Based on Fast Marching Method}

For human beings, after entering a completely unknown environment, with the help of instructions, they will obtain the information of nearby scenes through observation, and construct a general path to the destination in their mind, which can be corrected on the way, such as finding a closer way. While for the agent, because the basic information of the scene has been obtained through preliminary exploration, the language expectation converts the instructions into the task sequence that the agent can understand. Based on the dynamic 2D map $V_t$, the agent can quickly find the appropriate path of the target position $\mathcal{P}_t = \{p_{t_1},p_{t_2},...,p_{t_n}\} $ through the Fast Marching Method (FMM) \cite{FMM} according to its position $p_{r_t}$ and the position of the object to be operated $p_o$:
$$ \mathcal{P}_t = {\rm FMM}(p_{r_t},p_o,V_t)$$

In addition, in the navigation process, the agent will constantly update the map according to the existing situation and obtain the information of surrounding objects, which is conducive to the further implementation of subsequent tasks. Therefore, at each step $t$, the model will use the Fast Marching Method to obtain a new path to the target location.

\subsubsection{Determined Manipulation Policy}

For object manipulation, we design a corresponding determined set of atomic actions to execute a specific manipulation step. For example, After a navigation step to the microwave, the heating step always contains actions of toggling on and off the microwave, putting the object in, and then picking it up, opening the wave, and closing it. These standard step-by-step actions can be unified into action sets to execute sub-steps. %We design appropriate series of actions for agents corresponding to all manipulation types shown in table \ref{manipulation}. 
In this way, multiple simple actions are generalized as one determined policy, which we think can avoid unnecessary mistakes.

\begin{table*}[]
\renewcommand\arraystretch{1.3}
\centering
\scalebox{1.2}{
\begin{tabular}{clccccclclcllclcl}
\multicolumn{1}{l}{}                &                & \multicolumn{1}{l}{}   & \multicolumn{1}{l}{}   & \multicolumn{1}{l}{} & \multicolumn{1}{l}{}            & \multicolumn{1}{l}{}   &  & \multicolumn{1}{l}{}          &       & \multicolumn{1}{l}{}         &        &                      & \multicolumn{1}{l}{}          &       & \multicolumn{1}{l}{}          &       \\ \hline
\multicolumn{2}{c}{\multirow{4}{*}{\textbf{Method}}} & \multicolumn{5}{c}{\multirow{2}{*}{\textbf{Leaderboard Test Performance}}}                                                        &  & \multicolumn{9}{c}{\multirow{2}{*}{\textbf{Valid Performance}}}                                                                                                                     \\
\multicolumn{2}{c}{}                                 & \multicolumn{5}{c}{}                                                                                                              &  & \multicolumn{9}{c}{}                                                                                                                                                                 \\ \cline{3-4} \cline{6-7} \cline{9-12} \cline{14-17} 
\multicolumn{2}{c}{}                                 & \multicolumn{2}{c}{\textbf{Seen}}               & \multicolumn{1}{l}{} & \multicolumn{2}{c}{\textbf{Unseen}}                      &  & \multicolumn{4}{c}{\textbf{Seen}}                                             & \multicolumn{1}{c}{} & \multicolumn{4}{c}{\textbf{Unseen}}                                           \\ \cline{3-4} \cline{6-7} \cline{9-12} \cline{14-17} 
\multicolumn{2}{c}{}                                 & SR            & GC                     & \multicolumn{1}{l}{} & SR                              & GC                     &  & \multicolumn{2}{c}{SR}                & \multicolumn{2}{c}{GC}                & \multicolumn{1}{c}{} & \multicolumn{2}{c}{SR}                & \multicolumn{2}{c}{GC}                \\ \hline
\multicolumn{17}{l}{{\textit{\textbf{Step-by-Step Instructions + Goal Description}}}}   \\ \hline
\multicolumn{2}{l}{Seq2Seq \cite{shridhar2020alfred}}                          & 3.98                   & 9.42                   &                      & 0.39                            & 7.03                   &  & \multicolumn{2}{c}{3.70}              & \multicolumn{2}{c}{10.00}             &                      & \multicolumn{2}{c}{0.00}              & \multicolumn{2}{c}{6.90}              \\
\multicolumn{2}{l}{MOCA \cite{Moca}}                             & 22.05                  & 28.29                  &                      & 5.30                            & 14.28                  &  & \multicolumn{2}{c}{19.15}             & \multicolumn{2}{c}{28.5}              &                      & \multicolumn{2}{c}{3.78}              & \multicolumn{2}{c}{13.4}              \\
\multicolumn{2}{l}{ET \cite{ET}}                               & 38.42                  & 45.44                  &                      & 8.57                            & 18.56                  &  & \multicolumn{2}{c}{33.78}             & \multicolumn{2}{c}{42.48}             &                      & \multicolumn{2}{c}{3.17}              & \multicolumn{2}{c}{13.12}             \\
\multicolumn{2}{l}{LWIT \cite{LWIT}}                             & 30.92                  & 40.53                  &                      & 9.42                            & 20.91                  &  & \multicolumn{2}{c}{33.70}             & \multicolumn{2}{c}{43.10}             &                      & \multicolumn{2}{c}{9.70}              & \multicolumn{2}{c}{23.10}             \\
\multicolumn{2}{l}{HiTUT \cite{Hitut}}                            & 21.27                  & 29.97                  &                      & 13.87                           & 20.31                  &  & \multicolumn{2}{c}{25.24}             & \multicolumn{2}{c}{34.85}             &                      & \multicolumn{2}{c}{12.44}             & \multicolumn{2}{c}{23.71}             \\
\multicolumn{2}{l}{ABP \cite{ABP}}                              & \underline{44.55}            & \underline{51.13}            &                      & 15.43                           & 24.76                  &  & \multicolumn{2}{c}{\underline{42.93}}       & \multicolumn{2}{c}{\underline{50.45}}       &                      & \multicolumn{2}{c}{12.55}             & \multicolumn{2}{c}{25.19}             \\
\multicolumn{2}{l}{EmBERT \cite{Embert}}                           & 31.77                  & 39.27                  &                      & 7.52                            & 16.33                  &  & \multicolumn{2}{c}{37.44}             & \multicolumn{2}{c}{44.62}             & \multicolumn{1}{c}{} & \multicolumn{2}{c}{5.73}              & \multicolumn{2}{c}{15.91}             \\
\multicolumn{2}{l}{FILM \cite{Film}}                             & 27.67                  & 38.51                  &                      & \underline{26.49}                     & \underline{36.37}            &  & \multicolumn{2}{c}{24.63}             & \multicolumn{2}{c}{37.20}             &                      & \multicolumn{2}{c}{20.10}       & \multicolumn{2}{c}{\underline{32.45}}       \\ \hline
\multicolumn{17}{l}{{\textit{\textbf{Goal Description Only}}}}   \\ \hline
\multicolumn{2}{l}{HiTUT G \cite{Hitut}}                            & 18.41                  & 25.27                  &                      & 10.23                           & 20.27                  &  & \multicolumn{2}{c}{13.63}             & \multicolumn{2}{c}{21.11}             &                      & \multicolumn{2}{c}{11.12}             & \multicolumn{2}{c}{17.89}             \\
\multicolumn{2}{l}{LAV \cite{LAV}}                              & 13.35                  & 23.21                  &                      & 6.38                            & 17.27                  &  & \multicolumn{2}{c}{12.7}              & \multicolumn{2}{c}{23.4}              &                      & \multicolumn{2}{c}{-}                 & \multicolumn{2}{c}{-}                 \\
\multicolumn{2}{l}{HLSM \cite{HLSM}}                             & \textbf{29.94}         & \textbf{41.21}         &                      & 20.27                           & 30.31        &  & \multicolumn{2}{c}{\textbf{29.63}}    & \multicolumn{2}{c}{\textbf{38.74}}    &                      & \multicolumn{2}{c}{18.28}    & \multicolumn{2}{c}{\textbf{31.24}}    \\ \hline
\multicolumn{17}{l}{{\textit{\textbf{Step-by-Step Instructions Only}}}}  \\ \hline
\multicolumn{2}{l}{LEBP (ours)}                             & 25.53         & 32.35        &                      & \textbf{24.26}                        & \textbf{30.49}        &  & \multicolumn{2}{c}{27.63}    & \multicolumn{2}{c}{35.76}    &                      & \multicolumn{2}{c}{\underline{\textbf{22.36}}}    & \multicolumn{2}{c}{29.58}                 \\ \hline
\end{tabular}
}

\caption{The results on test seen/unseen and valid seen/unseen splits. The top section methods use both the step-by-step instructions and goal description. The middle section uses only goal description. The bottom section uses only step-by-step instructions, whereas our method is the unique one utilizing instructions only. Best results \textbf{using only single language guidance} (goal description or step-by-step instructions) and \underline{using both}  are highlighted.}
\vspace{-4em}
\label{Main}
\end{table*}

\begin{table}[]
\centering
\renewcommand\arraystretch{1.2}
\scalebox{1.1}{
\begin{tabular}{llcclcc}
\\ \hline
\multicolumn{2}{c}{\multirow{4}{*}{\textbf{Task Type}}} & \multicolumn{5}{c}{\multirow{2}{*}{\textbf{Validation}}}                                                                             \\
\multicolumn{2}{c}{}                                    & \multicolumn{5}{c}{}                                                                                                                 \\ \cline{3-4} \cline{6-7} 
\multicolumn{2}{c}{}                                    & \multicolumn{2}{c}{\textbf{Seen}}                     & \multicolumn{1}{l}{} & \multicolumn{2}{c}{\textbf{Unseen}}                   \\ \cline{3-4} \cline{6-7} 
\multicolumn{2}{c}{}                                    & SR               & GC                        & \multicolumn{1}{l}{} & SR                        & GC                        \\ \hline
\multicolumn{2}{l}{Overall}                             & \multicolumn{1}{l}{27.63} & \multicolumn{1}{l}{35.76} & \multicolumn{1}{l}{} & \multicolumn{1}{l}{22.36} & \multicolumn{1}{l}{29.58} \\ \hline
\multicolumn{2}{l}{Examine}                             & 45.24                     & 59.52                     &                      & 41.00                     & 58.08                     \\
\multicolumn{2}{l}{Pick \& Place}                       & 44.00                     & 44.00                     &                      & 22.81                     & 23.09                     \\
\multicolumn{2}{l}{Stack \& Place}                      & 24.76                     & 34.00                     &                      & 16.84                     & 20.13                     \\
\multicolumn{2}{l}{Clean \& Place}                      & 4.65                      & 25.19                     &                      & 0.00                      & 18.27                     \\
\multicolumn{2}{l}{Cool \& Place}                       & 12.4                      & 27.60                     &                      & 9.26                      & 22.22                     \\
\multicolumn{2}{l}{Heat \& Place}                       & 15.56                     & 31.53                     &                      & 3.81                      & 25.93                     \\
\multicolumn{2}{l}{Pick 2 \& Place}                     & 28.07                     & 36.47                     &                      & 11.25                     & 10.00                     \\ \hline
\multicolumn{1}{l}{}                 &                  & \multicolumn{1}{l}{}      & \multicolumn{1}{l}{}      & \multicolumn{1}{l}{} & \multicolumn{1}{l}{}      & \multicolumn{1}{l}{}     
\end{tabular}}
\vspace{-1.2em}
\caption{The performance in each task type on the validation splits.}
\vspace{-3em}
\label{Task}
\end{table}

\begin{table}[]
\centering
\renewcommand\arraystretch{1.2}
\scalebox{1.1}{
\begin{tabular}{llccccc}
                          &                          & \multicolumn{1}{l}{}     & \multicolumn{1}{l}{}     &                      & \multicolumn{1}{l}{}     & \multicolumn{1}{l}{}     \\ \hline
\multicolumn{2}{c}{\multirow{4}{*}{\textbf{Method}}} & \multicolumn{5}{c}{\multirow{2}{*}{\textbf{Validation}}}                                                                         \\
\multicolumn{2}{c}{}                                 & \multicolumn{5}{c}{}                                                                                                             \\ \cline{3-4} \cline{6-7} 
\multicolumn{2}{c}{}                                 & \multicolumn{2}{c}{\textbf{Seen}}                   &                      & \multicolumn{2}{c}{\textbf{Unseen}}                 \\ \cline{3-4} \cline{6-7} 
\multicolumn{2}{c}{}                                 & SR              & GC                       &                      & SR                       & GC                       \\ \hline
\multicolumn{2}{l}{LEBP}                             & 27.6                     & 35.8                     & \multicolumn{1}{c}{} & 22.4                     & 29.6                     \\ \hline
\multicolumn{2}{l}{+gt depth}                        & 29.8                     & 37.6                     & \multicolumn{1}{c}{} & 24.1                     & 32.4                     \\
\multicolumn{2}{l}{+gt depth, gt seg.}               & 51.1                     & 59.7                     & \multicolumn{1}{c}{} & 46.7                     & 50.8                     \\
\multicolumn{2}{l}{+gt seg.}                         & 45.6                     & 54.1                     & \multicolumn{1}{c}{} & 34.2                     & 42.4                     \\ \hline
\multicolumn{2}{l}{FILM}                             & 24.6                     & 37.2                     & \multicolumn{1}{c}{} & 20.1                     & 32.5                     \\ \hline
\multicolumn{2}{l}{+gt depth}                        & 26.6                     & 38.2                     & \multicolumn{1}{c}{} & 30.7                     & 42.9                     \\
\multicolumn{2}{l}{+gt depth, gt seg.}               & 43.2                     & 55.5                     & \multicolumn{1}{c}{} & 55.1                     & 64.3                     \\
\multicolumn{2}{l}{+gt seg.}                         & 34.0                     & 45.5                     & \multicolumn{1}{c}{} & 29.4                     & 42.9                     \\ \hline
\multicolumn{2}{l}{HLSM}                             & \multicolumn{1}{l}{29.6} & \multicolumn{1}{l}{38.8} &                      & \multicolumn{1}{l}{18.3} & \multicolumn{1}{l}{31.2} \\ \hline
\multicolumn{2}{l}{+gt depth}                        & 29.6                     & 40.5                     &                      & 20.1                     & 33.7                     \\
\multicolumn{2}{l}{+gt depth, gt seg.}               & 40.7                     & 50.4                     &                      & 40.2                     & 52.2                     \\
\multicolumn{2}{l}{+gt seg.}                         & 36.2                     & 47.0                     &                      & 34.7                     & 47.8                     \\ \hline
                          &                          & \multicolumn{1}{l}{}     & \multicolumn{1}{l}{}     &                      & \multicolumn{1}{l}{}     & \multicolumn{1}{l}{}     \\
                          &                          & \multicolumn{1}{l}{}     & \multicolumn{1}{l}{}     &                      & \multicolumn{1}{l}{}     & \multicolumn{1}{l}{}    
\end{tabular}}
\vspace{-2.8em}
\caption{Ablation results on validation splits.}
\vspace{-3.5em}
\label{Ablation}
\end{table}

\section{Experiments and Results}

To evaluate LEBP, we conduct experiments on the ALFRED benchmark. Compared with other methods, our method does not use additional data (e.g., additional language instructions \cite{ET} and visual distortion \cite{ABP}).

\subsection{Experimental Setup} 

\subsubsection{Environment \& Dataset} The ALFRED benchmark \cite{shridhar2020alfred} combines household task demonstrations with language instructions in 3D simulated rooms \cite{ai2thor} and consists of seven task types: Pick \& Place, Stack \& Place, Pick 2 \& Place, Heat \& Place, Cool \& Place, Clean \& Place, Examine. It contains 21,023 train, 1,642 validation, and 3,062 test tasks. The validation set and test set include Seen and Unseen scenarios, in which Seen scenario is included in the training set, \textbf{while the Unseen scenario for evaluation is an unknown new environment}. There are 8,055 expert demonstrations and 25,743 language instructions in the dataset. The expert demonstration includes the observations and actions at each step, and the language instructions mainly include high-level goal instructions and low-level step-by-step instructions, e.g. for the goal instruction \emph{examine the book under the light}, low-level step-by-step instructions will guide the agent how to get the book, pick up the book, go to the FloorLamp, turn on the light and examine it.

\subsubsection{Metrics} Success Rate (SR) is the fraction of whether all sub-goal was completed. The goal-condition success (GC) is the fraction of goal conditions completed across all tasks. 

\subsubsection{Implementation Details} The experiments are conducted on a server with a Dual 12-core 3.2GHz Intel Xeon and utilized 8 NVIDIA Tesla V100 with 32GB running on CUDA version 10.1. LEBP is implemented using PyTorch \cite{paszke2019pytorch}, skfmm, scikit-learn \cite{pedregosa2011scikit}, Huggingface-Transformers \cite{wolf2019huggingface}. For the language expectation module, we trained for 200 epochs with a batch size of 16 and We used the Adam optimizer with a learning rate of $5e^{-5}$. For the depth module, we trained for 100 epochs with a batch size of 32 and learning rate of $1e^{-4}$ and the semantic segmentation module, we trained for 250 epochs with a batch size of 32 and learning rate of $2e^{-1}$. After training, we fix the weight of the above modules, and then let the agent execute the tasks.

\subsection{Results}

Table \ref{Main} shows test and validation results. LEBP achieves competitive performance across both seen and unseen environments in the setting. The primary leaderboard metric is the unseen SR, measuring the model’s generalization abilities. LEBP achieves 3.99\% absolute (19.68\% relative) gain in SR, 0.18\% absolute (0.59\% relative) gain in GC on Test Unseen, and 4.08\% absolute (22.31\% relative) gain in SR on Valid Unseen over HLSM, which shows that under the condition of using only single language guidance, our performance is the best.

Our approach performs competitively even compared to methods that require both step-by-step instructions and goal description. Compared to FILM, we achieve 2.26\% absolute (12.44\% relative) improvement in SR on Valid Unseen. Moreover, the performance of LEBP on Unseen split is close to that of Seen split, which indicates that it can well adapt to unknown scenarios. Notice that among the end-to-end style methods, such as ABP \cite{ABP}, LWIT \cite{LWIT} and EmBERT \cite{Embert}, the score on Seen split is much higher than that on Unseen split, indicating risks of overfitting to the Seen split.

Table \ref{Task} show development results. To analyze the advantages and disadvantages of LEBP in different task types, we divide the results on the validation splits according to task types. We can see that our approach performs well on the “Examine”, with high SR and GC. However, for "Clean \& Place", "Cool \& Place" and "Heat \& Place", GC is much higher than SR, which indicates that there may often be a subtask that fails, failing in the whole task.

To confirm the impact of the visual module on our approach, we consider ablation on the base model, with ground truth visual input. Table \ref{Ablation} shows ablations on the validation splits. Ground truth segmentation (+gt seg.) provides 18.0\%/11.8\% absolute improvement in Seen/Unseen scenes, while ground truth depth (+gt depth) provides 1.8\%/1.7\% absolute improvement in Seen/Unseen scenes, which indicates that segmentation is the main bottleneck affecting the performance. Compared with the ablation results of other models, the performance of LEBP is competitive, showing the superiority of the language expectation module.

%The main failure of LEBP is due to 1) the wrong prediction action type or object type caused by language expectation error, 2) the object cannot be found with insufficient exploration, 3) the object to be manipulated is in a closed container, 4) the manipulation fails due to vision module error, 5) the appropriate path cannot be found for a long time during navigation, and 6) others. Specifically, finding and interacting with objects may be a major problem, shown from the model with ground truth segmentation result in table \ref{Ablation} that more active and accurate exploration strategies may be needed in the future.

\section{Conclusion}

We proposed a method Language Expectation \& Binding Policy (LEBP), a well-designed two-stream framework to separate the language understanding and action execution in an embodied agent where (1) contains a language expectation module, generating expectation to describe how the agent will execute tasks in sub-steps based on the prior knowledge of the scenario. (2) Binding policy addresses sub-steps in generated expectation into actual actions including navigation and object manipulation sub-step. Experimental results show that the ALFRED benchmark, LEBP achieves comparable performance to currently published SOTA methods. Our method provides a two-stream framework that can be adopted by ALFRED and beyond, and has better language interaction insight with humans via expectation. We believe our work provides inspiration and concepts for future embodied studies.

\bibliographystyle{apalike}
\bibliography{ref}

\end{document}